\pdfoutput=1
\documentclass[11pt]{article}

\usepackage{acl2023}

\usepackage{times}
\usepackage{latexsym}
\usepackage{graphicx}
\usepackage{pgfplots}
\usepackage{multirow}
\usepackage{float}
\usepackage{inconsolata}
\usepackage{booktabs}
\usepackage{amsmath}
\usepackage{microtype}
\usepackage{rotating}
\usepackage{graphicx}
\usepackage{amsfonts}
\usepackage{booktabs}
\usepackage{color}
\usepackage{comment}
\usepackage{amsmath,amsfonts,amssymb}
\usepackage{arydshln}
\usepackage{tablefootnote}
\usepackage{xcolor}
\usepackage{makecell}
\usepackage{hyperref}
\usepackage{pifont}% http://ctan.org/pkg/pifont

\usepackage[T1]{fontenc}
\usepackage[utf8]{inputenc}

\usepackage{microtype}

\title{Tab-CoT: Zero-shot Tabular Chain of Thought}

\author{Ziqi Jin \and Wei Lu\\
    StatNLP Research Group\\
    Singapore University of Technology and Design \\
    \texttt{ziqi\_jin@sutd.edu.sg, luwei@sutd.edu.sg} \\}
    \pgfplotsset{compat=1.18}
\begin{document}
\maketitle
\begin{abstract}
% Todo: CoT,
The chain-of-though (CoT) prompting methods were successful in various natural language processing (NLP) tasks thanks to their ability to unveil the underlying complex reasoning processes.
Such reasoning processes typically exhibit implicitly structured steps.
Recent efforts also started investigating methods to encourage more explicitly structured reasoning procedures to be captured \cite{https://doi.org/10.48550/arxiv.2205.10625}.
In this work, we propose Tab-CoT, a novel tabular-format CoT prompting method, which allows the complex reasoning process to be explicitly modelled in a highly structured manner.
Despite its simplicity, we show that our approach is capable of performing reasoning across multiple dimensions (i.e., both rows and columns).
We demonstrate our approach's strong zero-shot and few-shot capabilities through extensive experiments on a range of reasoning tasks.\footnote{Our code is available at \url{https://github.com/Xalp/Tab-CoT}}

%Current Chain-of-Thought (CoT) methods are making significant improvements in terms of reasoning abilities.
%However, natural language contains many structural connections and grammar rules, which could be redundant for the reasoning process.
%We propose Tab-CoT, a tabular-format CoT method that produces a more organized and clean table as processes.
%While the language itself is 1-dimensional, our result shows that the table can flow information from 2 dimensions (columns and rows). This helps our method conduct 2-dimensional reasoning. 
%Different from other methods, our method can be used in both zero-shot and few-shot settings. It outperformed similar work in 5 out of 6 arithmetic reasoning tasks and gain highly competitive results in the few-shot setting. 
%\footnote{Our code is available at *URL*}
\end{abstract}

\section{Introduction}

% content to include:
% CoT; v
% Zero-shot-CoT; v
% "code-davinci-002" --> Table; v
% Table is computationally heavy/organized / ...
% Our propose
% Experiment result

The chain-of-thought (CoT) prompting method \cite{https://doi.org/10.48550/arxiv.2201.11903} encourages the large language models (LLMs) to engage in a thought process before providing the answer to the given question.
Such an approach shows impressive performance improvements in reasoning tasks.
Notably, in the zero-shot setting, it was shown that a simple prompt such as ``\texttt{let's think step by step}'' could facilitate the step-by-step thinking process before answering the original question \cite{llmrzr}.
Such a task-agnostic method unveiled that LLMs can be descent zero-shot reasoners.

\begin{figure}[t!]
\centering
    \includegraphics[scale=0.50]{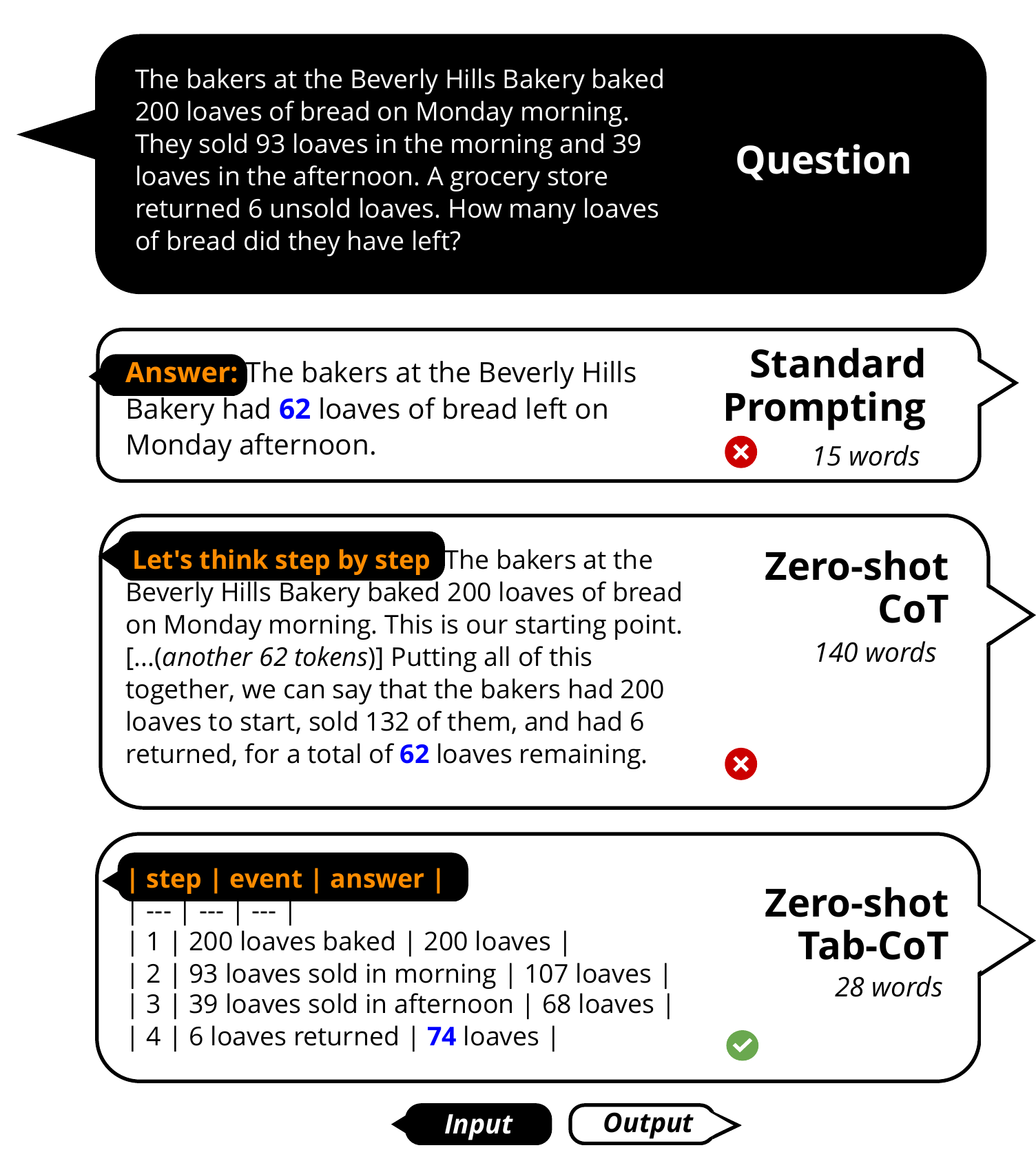}
    \caption{A comparison between Tab-CoT with standard prompting and zero-shot-CoT on the same question. Chain-of-thought prompts are highlighted in \textcolor{orange}{orange}.}
    \vspace{-7mm}
    \label{fig:example_all}
\end{figure}

The reasoning process is inherently structured. This gives rise to some new developments along this line of work recently. Specifically, \citet{https://doi.org/10.48550/arxiv.2205.10625} suggests an alternative prompting approach that enables a two-stage structured reasoning process.
\citet{https://doi.org/10.48550/arxiv.2211.10435} proposes an approach that involves code in the prompt design, allowing structured information in the form of formal language to participate in the reasoning process.
While effective, such methods require specific prompt engineering for different domains or defining multiple variables, which can be difficult to maintain or keep track of.

%However, the current chain of thought is restricted by the language rules. While previous work mentioned that code could replace language for reasoning steps \cite{https://doi.org/10.48550/arxiv.2211.10435}, we found that the code does not show intermediate values like the table does.
%Code contains a lot of variables, and their values will not be shown in the code (even if you use the print function, the result is not in the code). The large language model needs to keep track of multiple variables and perform calculations. This way, the table is much more explicit as all values will be shown. Moreover, the table does not contain fixed functions and rules as coding language does.

Inspired by the fact that state-of-the-art large language models, such as GPT-3 \cite{https://doi.org/10.48550/arxiv.2005.14165} and CodeX \cite{https://doi.org/10.48550/arxiv.2107.03374}, have the capability of reasoning over tabular structured data \footnote{This is because such models are trained on massive data collected from the Internet, which contains a large amount of tabular formed data.}, we propose a novel framework called {\em Tabular Chain of Thought} (Tab-CoT) that models the structured reasoning process using a table-filling procedure.

We show that the model can perform step-by-step reasoning by creating a table without further fine-tuning by using a table header with column names in the form of ``\texttt{|step|question|response|}'' as a prompt.
While conventional natural language texts are generated in a 1-dimensional sequential order, the table has a 2-dimensional structure, allowing inference along both columns and rows to be performed simultaneously.
Unlike previous works which focused on extracting information from existing tabular structured data (\citealp{gong-etal-2020-tablegpt}), our approach generates the table while performing the reasoning process (and extracts the answer from the generated table at the end).
%We named this ability of reasoning from both directions as ``2-dimensional reasoning''.

\begin{figure*}[t!]
\centering
    \includegraphics[scale=0.6]{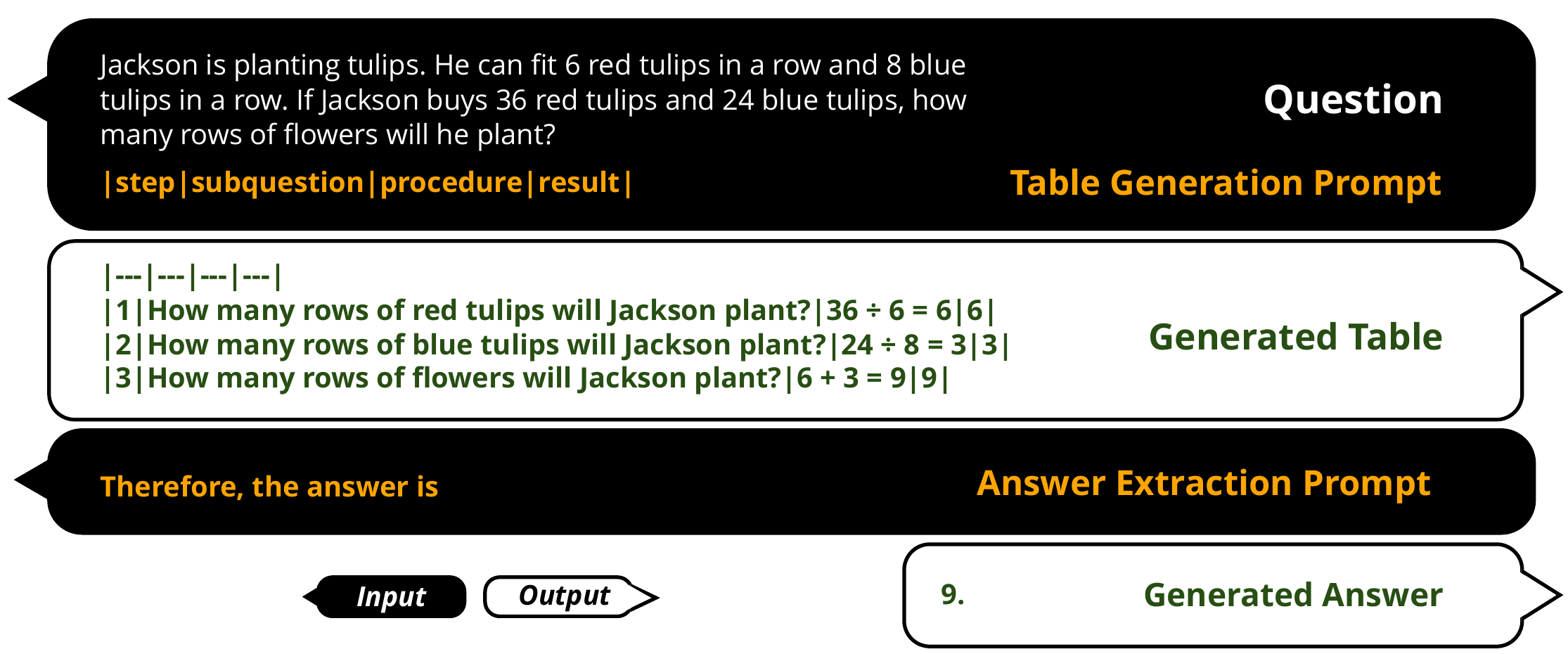}
    \vspace{-2mm}
    \caption{Overview of our zero-shot Tab-CoT method, which contains two steps: (1) table generation and (2) answer extraction. Added prompts are highlighted in \textcolor{orange}{orange}. Texts generated by the LLM are highlighted in \textcolor{teal}{green}.}
    % \vspace{-3mm}
    \label{fig:tab_cot}
\end{figure*}

Figure \ref{fig:example_all} shows the results with standard prompting, conventional zero-shot CoT, and our zero-shot Tab-CoT.
Our method generates a table as the output, which is more organized and concise than the output from the conventional CoT method.
In this example, while zero-shot CoT generates 140 words, our method only generates 28.
Besides, we found our method can reason horizontally and vertically at the same time.\footnote{ ``\texttt{74 loaves}'' is the sum of  ``\texttt{68 loaves}'' from the same row and ``\texttt{6 loaves}'' from the same column.}
This demonstrates that our Tab-CoT method benefits from the 2-dimensional structure of the table, where the information can flow in two dimensions.

We summarize our main contributions in this work as follows:
\begin{itemize}
    \setlength\itemsep{1em}
  \item We propose a new approach called Tabular Chain-of-Thought (Tab-CoT) that utilizes a tabular structured reasoning scheme in combination with state-of-the-art large language models to generate answers. To the best of our knowledge, this is the first method that uses tables in a ``chain of thought'' process.
  \item The 2-dimensional tabular structure of Tab-CoT allows for improved unlocking of the step-by-step reasoning capabilities of LLMs, transforming the linear ``chain of thought'' process into a more structured one.
  \item Extensive experiments have revealed that our Tab-CoT outperforms traditional CoT techniques in zero and few-shot settings. This indicates that Tab-CoT has strong potential as a superior alternative to current chain-of-thought prompting methods.
\end{itemize}

% method figure

\section{Related Work}
% CoT
Chain-of-thought prompting \cite{https://doi.org/10.48550/arxiv.2201.11903}, a variation of few-shot prompting that adds step-by-step reasoning in those few-shot examples instead of just providing answers, has achieved significant improvements across multiple datasets.
The LLMs can generate solutions following the solution format of prompts. Compared to traditional prompting, chain-of-thought prompting decomposes the task into smaller steps, which makes difficult tasks easier to solve.

The chain-of-thought prompting method is not necessarily purely natural language based. Program Aided Language Models (PAL) \cite{https://doi.org/10.48550/arxiv.2211.10435} provides few-shot samples that contain executable Python code.
Such an approach enables the LLMs to interact with the Python shell, allowing the model to focus on learning how to do mathematical reasoning rather than numerical calculations.

These chain-of-thought methods provide the solution structure and pattern via few-shot samples, but can these be provided without these few-shot samples in the zero-shot setting?
Zero-shot CoT \cite{llmrzr} is a zero-shot chain-of-thought prompting method. The prompt phrase ``\texttt{Let's think step by step}'' added after the question triggers the explicit reasoning process. 
However, compared to few-shot CoT \cite{https://doi.org/10.48550/arxiv.2201.11903}, zero-shot CoT allows more flexibility in the structure of the reasoning process.

Recently, \citet{https://doi.org/10.48550/arxiv.2205.10625} proposed Least-to-Most prompting, which is a prompting strategy that reduces a complex problem into a list of sub-questions, and sequentially solves the sub-questions. Each sub-question is solved with the answer to previously solved sub-questions. Compared to zero-shot CoT, this method has more restrictions on the structure of reasoning by decomposing and sequentially answering. Moreover, importing external tools (like calculator and python shell) can further aid the math computation within the arithmetic domain \cite{https://doi.org/10.48550/arxiv.2211.10435}.

These works reveal the importance of promoting structures in the chain-of-thought process.
However, the nature of the zero-shot prompting makes the injection of structures into the generation process challenging.
This motivates us to devise a better mechanism to prompt the language models under the zero-shot setting -- a new prompting scheme that allows highly structured outputs in the form of tables to be generated.

\begin{figure*}[t!]
\centering
 \vspace{-2mm}
    \includegraphics[scale=0.45]{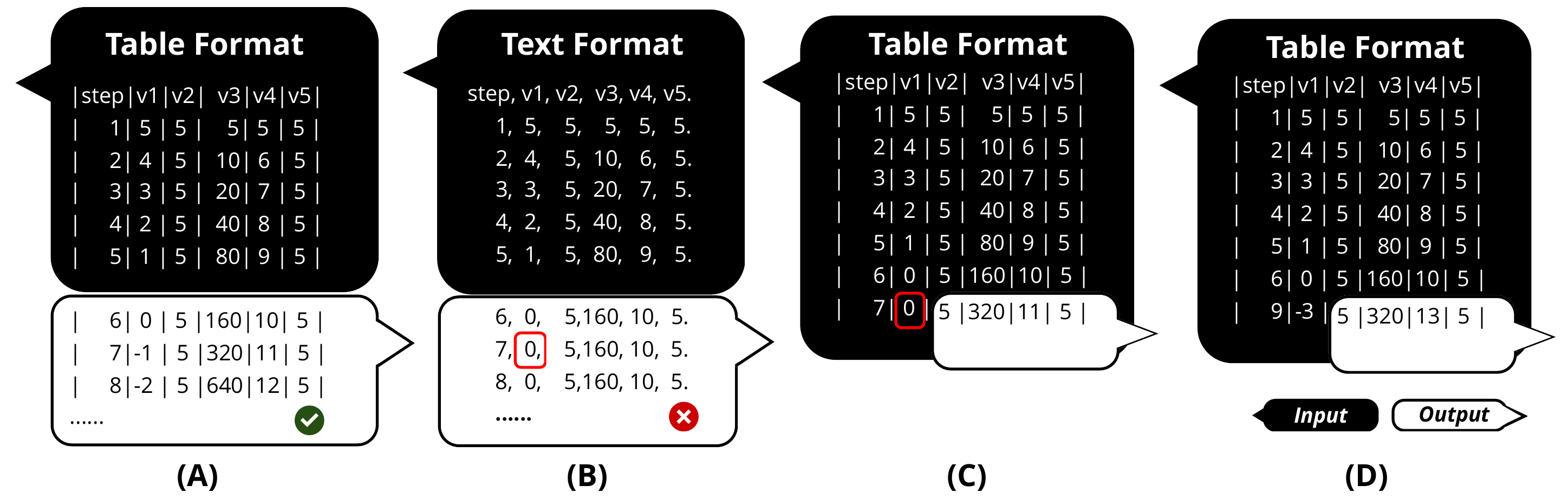}
     \vspace{-2mm}
    \caption{Understanding how state-of-the-art LLM (``code-davinci-002'') reason with tabular-structured data.} % [JZQ] TABLE CONTENT JUZHONG
    % \vspace{-4mm}
    \label{fig:exp_table}
\end{figure*}

\section{Tab-CoT}

Similar to zero-shot CoT \cite{llmrzr}, our method involves two prompts that can be used in large language models (LLMs), one for table generation and the other for answer extraction.
The details are shown in Figure \ref{fig:tab_cot}. While our method is primarily applied in zero-shot settings, it can also work in few-shot settings.

% todo: add how you discover [block experiment]
\paragraph{Tables in LLMs}

% As we believe we can use LLMs to perform structured reasoning with tabular-structured data, the first key question we would like to ask is how the state-of-the-art LLMs handle such 2-dimensional structured representations.

We found that in the official ``parse unstructured data'' demo provided by OpenAI\footnote{\url{https://beta.openai.com/playground/p/default-parse-data}}, a table header is provided as part of the prompt, which is as follows: ``\texttt{|Fruit|Color|Flavor|}''. With such a prompt, the underlying LLM can automatically generate a table.
This suggests possible formatting for tables in such state-of-the-art LLMs.
And ``|'' is the recognizable delimiter of tables in OpenAI models.

To validate this observation, we queried the LLM ``code-davinci-002'' \cite{https://doi.org/10.48550/arxiv.2107.03374} with the following question: ``\texttt{a=2, b=3, what is 2*a+3*b?}'', and provided another table header: ``\texttt{|step|solution|}''\footnote{The temperature is set to 0 for reproducibility.}.
We found that it completes a structured table as follows:

\begin{small}
\begin{verbatim}
            a=2, b=3, what is 2*a+3*b?
            |step|solution|
            |:---|:---|
            |1|2*a+3*b|
            |2|2*2+3*3|
            |3|4+9|
            |4|13|
\end{verbatim}
\end{small}

This experiment essentially unveils how the tables are represented in such LLMs. The results also illustrate how the table can potentially be used for generating a reasoning process. Next, to validate this, we designed several simple experiments to understand how reasoning over such tabular-structured data is performed on such LLMs, as shown in Figure \ref{fig:exp_table}.
Our first experiment (A) shows that such LLMs are able to perform potential vertical reasoning.
However, if we replace `\texttt{|}' with `\texttt{,}' (B), the LLM fails to capture the patterns in the data.
This tells us that the correct formatting is crucial when reasoning with tables in such LLMs.

Next, we intentionally insert a mistake into the partial table and ask the model to continue the generation process (circled in C). Surprisingly, the LLM is able to generate the correct entries even though the mistake occurred in the same row. This further confirms the LLM's strong potential in performing vertical reasoning with tabular-structured data.

Moreover, to prove both vertical and horizontal reasoning exists, we increase the difficulty by directly appending the first two elements from step 9 after step 6 (D).
If only vertical reasoning existed, the value under ``\texttt{v4}'' would have been ``\texttt{11}''.
Instead, the value generated is ``\texttt{13},'' confirming that the LLMs have the potential to perform a combination of horizontal and vertical reasoning simultaneously.

\paragraph{Table Generation Prompt} To make use of the 2-dimensional structure of the table, we replace the natural language prompt with a table-generation prompt (e.g., ``\texttt{|step|question|response|}''), which serves as the header of the table.
This regulates the context of this table, forcing the LLMs to conduct step-by-step reasoning by completing the table. Meanwhile, the choice of columns can be very specific. 
If each row of the table is regarded as a step, the row-by-row table generation process will become a step-by-step reasoning process. Within each step (row), we have multiple columns, each of which contributes certain detail towards the current reasoning step. 

For any text question \(x\), we have a table generation prompt (all column names) \(c\). Concretely, we add the table generation prompt in the next row of the text question:

\vspace{-1.8em}
\begin{equation}
    \textsf{LLM}(x,c)=  \begin{bmatrix}
                c_{1} &  c_{2} & \cdots & c_{n-1} & c_{n}\\ 
                t_{1,1} &  t_{1,2} & \cdots & t_{1,n-1} & t_{1,n} \\ 
                \vdots &  & \ddots &  & \vdots\\ 
                t_{m,1} &  t_{m,2} & \cdots & t_{m,n-1} & t_{m,n}
                \end{bmatrix}
    % \vspace{-0.5em}
\end{equation}
where \(t_{1,1} \cdots t_{m,n}\) are the entries within the generated table, which contains \(m\) rows and \(n\) columns.

\paragraph{Answer Extraction Prompt} After the table content, denoted as \(T\), is generated from the previous step,
we perform answer extraction.
The answer extraction step helps us to extract the answer from the table, as the final results may not always be in the last cell of the generated table.
Following zero-shot CoT \cite{llmrzr}, we add another answer extraction prompt \(a\): ``\texttt{the answer is}'' after the generated table,
 to {extract} the final answer from the table:
\begin{equation}
   {\mathit Answer} =  \textsf{LLM} ( x,c,T,a)
\end{equation}

\paragraph{Structure-Promoting Table Scheme} 
Different table generation prompts (headers) may result in different tables generated (with different content).
We propose a ``structure-promoting scheme'', which maximally unlocks the reasoning abilities of LLMs.

We define each row as a reasoning step.
A table containing multiple rows will depict the step-by-step reasoning procedure leading to the final answer.
Thus, our first column is ``\texttt{step}'', containing a number that indicates which reasoning step the current row represents.

Least-to-most prompting \cite{https://doi.org/10.48550/arxiv.2205.10625} contains two stages: problem reduction and sequential solving.
In problem reduction, they decompose a question into multiple subquestions.
Similarly, we add ``\texttt{subquestion}'' as our second column.
At the beginning of each step, the LLMs will generate a subquestion under this column, which demonstrates the objective of the current reasoning step.

The conventional zero-shot CoT \cite{llmrzr} shows that allowing the model to generate some reasoning process before answering can achieve a better result.
Inspired by this observation, we add a third column, ``\texttt{process}'', into our table.
Given a subquestion in the previous column, we expect to generate the reasoning process in the current column before answering.

The last column is named ``\texttt{answer}''.
As the previous reasoning process under the ``\texttt{process}'' column may not necessarily provide an answer, we hope to use the ``\texttt{answer}'' column to explicitly request an (intermediate) answer at the end of each reasoning step.

With the above considerations, our primary scheme for the table header is designed as follows, which serves as our main table generation prompt:
\begin{verbatim}
   |step|subquestion|process|result|
\end{verbatim}

\begin{table}[t!]
		\centering
		\scalebox{0.68}{
			\begin{tabular}{llccc}
				\toprule
                & Reasoning Type & Dataset & Size & Answer Type\\
				\midrule
                \multirow{6}*{\hspace*{0.6em}\turnbox{90}{\hspace*{-.6em}Main} } & \multirow{6}*{{Arithmetic}}
				& SingleEq & \textcolor{white}{0,}508 & Numeral\\
				&& AddSub & \textcolor{white}{0,}395 & Numeral\\
				&& MultiArith & \textcolor{white}{0,}600 & Numeral\\
				&& GSM8K & 1,319 & Numeral\\
				&& AQUA & \textcolor{white}{0,}254 & Multiple Choice\\
				&& SVAMP & 1,000 & Numeral\\
				\midrule
                \multirow{4}*{\hspace*{0.6em}\turnbox{90}{\hspace*{-2em}Additional} } &
				\multirow{2}*{Symbolic}
				& Coin Flip & 1,000 & Yes or No\\
				&& Last Letter & \textcolor{white}{0,}254 & String\\
                \cmidrule{2-5}
				&\multirow{2}*{Commonsense}
				& StrategyQA & 2,290 & Yes or No\\
				&& CommonsenseQA & 1,221 & Multiple Choice\\
				\bottomrule
			\end{tabular}
		}
		\caption{Tasks and Data}
		\vspace{-4mm}
  \label{tab:data}
	\end{table}
 
% Tabular --> give column;
% arithmetic
% main exp table
\begin{table*}[t!]
		\centering
		\scalebox{0.62}{
			\begin{tabular}{llcccccccc|c}
				\toprule
                & Method & CoT Prompt & LLM & SingleEq & AddSub & MultiArith & GSM8K & AQUA & SVAMP & Average\\
				\midrule
                \multirow{6}*{\hspace*{0.6em}\turnbox{90}{\hspace*{-2em}Zero-shot} } &
				\multirow{2}*{Standard Prompting} & \multirow{2}*{---} 
				& text & 74.6 & 72.2 & 17.7 & 10.4 & 22.4 & 58.8 & 42.7\\
                %\cmidrule{3-10}
				&&& code & 46.3 & 51.4 & \textcolor{white}{0}7.2 & \textcolor{white}{0}4.1 & 23.6 & 29.5 & 27.0\\
				\cmidrule{2-11}
				&\multirow{2}*{CoT} & \multirow{2}*{{\texttt{Let's think step by step}}}
				& text & 78.0 & 69.6 & 78.7 & 40.7 & 33.5 & \textbf{62.1} & 60.4\\
                %\cmidrule{3-10}
				&&& code & 65.6 & 65.6 & 64.8 & 31.8 & 29.5 & 39.9 & 49.5\\
				\cmidrule{2-11}
				&\multirow{2}*{Tab-CoT}
    %             & \multirow{2}*{|step|question|response|} 
				% & code & 77.6 & 73.9 & 79.0 & 38.0 & 34.3 & \textbf{63.9} & 61.2\\
    %             \cmidrule{3-10}
				% && text & \textbf{80.1} & \textbf{74.9} & 62.2 & 32.2 & 31.9 & 61.6 & 57.1\\
				& \multirow{2}*{{\texttt{|step|subquestion|process|result|}}} 
				& text & 74.6 & \textbf{71.9} & 72.2 & 39.3 & 36.6 & 57.0 & 58.6\\
    %\cmidrule{3-10}
				&&& code & \textbf{81.9} & 70.9 & \textbf{81.2} & \textbf{44.4} & \textbf{37.0} & 60.5 & \textbf{62.6}\\ 
				% & \multirow{2}*{|step|subquestion|procedure|result|} 
				% & \textbf{code} & \textbf{83.7} & 69.1 & 77.8 & 43.4 & 38.2 & 60.4 & 62.1\\
				% && text & 75.8 & 72.9 & 68.5 & 37.5 & \textbf{39.0} & 58.0 & 58.6\\
				\bottomrule
			\end{tabular}
		}
         \vspace{-2mm}
		\caption{Zero-shot results on the arithmetic datasets. All methods use the same answer extraction prompt in these datasets for a fair comparison. All methods are evaluated under the zero-shot setting.}
		\vspace{-4mm}
  \label{tab:results_zero_arith}
	\end{table*}

\section{Experimental Setup}

\paragraph{Large Language Models}

We consider two state-of-the-art large language models under the GPT-3 family \cite{https://doi.org/10.48550/arxiv.2005.14165} in our experiments, namely ``code-davinci-002'' and ``text-davinci-002'', whose APIs are made available by OpenAI\footnote{\url{https://openai.com/api/}}.
For brevity, we use ``code'' to refer to the model ``code-davinci-002'' and ``text'' to refer to ``text-davinci-002'' in our experiments.

\paragraph{Tasks and Datasets}

We primarily focus on mathematical reasoning in this work. Thus, we evaluate our method on 6 arithmetic reasoning datasets.
Specifically, they are SingleEq \cite{koncel-kedziorski-etal-2015-parsing}, AddSub \cite{hosseini-etal-2014-learning}, MultiArith \cite{roy-roth-2015-solving}, GSM8K \cite{https://doi.org/10.48550/arxiv.2110.14168}, AQUA-RAT \cite{ling-etal-2017-program}, and SVAMP \cite{patel-etal-2021-nlp}, which are standard datasets widely used in the community.

We also conducted additional experiments on datasets involving other types of reasoning tasks.
Specifically, we evaluate our method on two symbolic reasoning tasks: Last letter and Coin Flip\footnote{We use the file generated by \citet{llmrzr}.}: the former is the task that asks for the concatenation of the last letters of 4 words, and the latter asks for the state of the coin after being flipped a few times.
We investigate how the specificity of column names affects the performance and report in our ablation study.
We also evaluate our method on two commonsense reasoning tasks: CommonsenseQA \cite{talmor-etal-2019-commonsenseqa} and StrategyQA \cite{geva-etal-2021-aristotle}.

Following zero-shot CoT \cite{llmrzr}, we set the first generated number as the numeral answer, the first capitalized letter as the answer for multiple-choice questions, and the first ``yes'' or ``no'' as the answer for ``Yes or No'' questions.

\section{Results}

\subsection{Main Results}

Our main experiments are conducted on arithmetic reasoning tasks under the zero-shot setting.
We tested the performance of both text-based and code-based LLMs on all methods.
The results are shown in Table \ref{tab:results_zero_arith}.
%All schemes achieve higher average accuracy than the conventional CoT under the zero-shot setting.
Under the scheme ``{\texttt{|step|subquestion|process|result|}}'', our zero-shot Tab-CoT approach significantly outperformed the standard prompting in all tasks.
Furthermore, our best-performing Tab-CoT model (using code-based LLM) outperforms the best conventional CoT model in 5 out of 6 tasks (with an average improvement of 2.2\%).
 
% \begin{table*}[t!]
% 		\centering
% 		\scalebox{0.62}{
% 			\begin{tabular}{llcccccc | c}
% 				\toprule
%                & Method & SingleEq & AddSub & MultiArith & GSM8K & AQUA & SVAMP & Average\\
% 				\midrule
%                 \multirow{3}*{\scalebox{1}{\hspace*{0.6em}\turnbox{90}{\hspace*{-1.6em}Few-shot} }} &
% %                Zero-shot Standard Prompting & 46.3 & 51.4 & \textcolor{white}{0}7.2 & \textcolor{white}{0}4.1 & 23.6 & 29.5 & 27.0\\
% %				Zero-shot-CoT & 65.6 & 65.6 & 64.8 & 31.8 & 29.5 & 39.9 & 49.5\\
% %				Zero-shot Tab-CoT & 81.9 & 70.9 & 81.2 & 44.4 & 37.0 & 60.5 & 62.6\\
% %				Zero-shot Tab-CoT + Majority Voting (3) & 86.4 & 78.2 & 85.2 & 48.2 & 44.1 & 66.9 & 68.2\\
% %				\midrule
% 				Standard Prompting & 86.8 & {\bf 90.9} & 44.0 & 19.7 & 29.5 & 69.9 & 68.2\\
% 				&CoT& {\bf 93.1} & 89.1 & 96.2 & {\bf 63.1} & 45.3 & 76.4 & 77.2\\
% 				&Tab-CoT & 92.1 & 89.1 & {\bf 96.3} & 61.6 & {\bf 46.9} & {\bf 82.9} & \textbf{78.2}\\
% 				\bottomrule
% 			\end{tabular}
% 		}
% 		\caption{Few-shot results on the arithmetic datasets. }
%   \label{tab:results_fewshot}
% 	\end{table*}

When the standard prompting method is considered, using the text-based LLM leads to significantly better results than the code-based counterpart (15.7\% on average).
Similarly, when zero-shot CoT is considered, using the former also outperforms the latter by 10.9\% on average.
However, for our Tab-CoT approach, ``code'' outperforms ``text'' by 4.0\%, leading to the best overall performance among all configurations.

From such results, we can see that the conventional CoT method responds differently from our Tab-CoT method with different types of underlying LLMs involved.
The conventional CoT method (and the standard prompting method) strongly favors a text-based LLM under the zero-shot setting.
In contrast, our approach works well with both types of LLMs, but the code-based version can give it an additional boost in performance.
Compared with ``text'', the ``code'' model is further fine-tuned on code \cite{https://doi.org/10.48550/arxiv.2107.03374}. 
We conjecture that table generation resembles the code generation process -- both involve structured procedures that are highly organized and follow a step-by-step process. Comparing our Tab-CoT approach with conventional CoT, we can conclude that our proposed table-generation prompt is able to significantly better unlock the strong reasoning abilities within the code-based LLM.

Based on the above main experiments, we choose to use ``code'' as the default LLM for all subsequent experiments unless otherwise specified.
 
\begin{table*}[t!]
		\centering
		\scalebox{0.62}{
			\begin{tabular}{llcccccc|c}
				\toprule
                & Scheme & SingleEq & AddSub & MultiArith & GSM8K & AQUA & SVAMP & Average\\
                \midrule
                \multirow{5}*{\hspace*{0.6em}\turnbox{90}{\hspace*{-2em}Zero-shot} } &
                Standard Prompting & 46.3 &	51.4&7.2&4.1&23.6&29.5 & 27.0 \\
				\cmidrule{2-9}
				&{\texttt{|step|subquestion|process|result|}} & 81.9&70.9&81.2&44.4&37.0&60.5  &62.6\\
				&{\texttt{|step|subquestion|procedure|result|}} & 83.7&69.1&77.8&43.4&38.2&60.4 & 62.1\\
				&{\texttt{|step|question|response|}}&77.6&73.9&79.0&38.1&34.3&63.9&61.1\\
				&Self-consistency (using above) & {\bf 86.4} & {\bf 78.2} & {\bf 85.2} & {\bf 48.2} & {\bf 44.1}& {\bf 66.9} & {\bf 68.2}\\
				\bottomrule
			\end{tabular}
		}
		\caption{Zero-shot performance comparison between the three schemes (and with self-consistency).}
		\vspace{-2mm}
  \label{tab:results_vote}
	\end{table*}
 
 \begin{table}[t!]
		\centering
		\scalebox{0.62}{
			\begin{tabular}{llc}
				\toprule
                & Scheme & Average\\
                \midrule
                \multirow{5}*{\hspace*{0.6em}\turnbox{90}{\hspace*{-2em}Zero-shot} } &
				{\texttt{|subquestion|process|result|}} 	&54.3\\
				&{\texttt{|step|process|result|}} 	&57.2\\
				&{\texttt{|step|subquestion|result|}}&61.3\\
				&{\texttt{|step|subquestion|process|}} 	&60.9\\
				\cmidrule{2-3}
				&{\texttt{|step|subquestion|process|result|}} &{\bf 62.6}\\
				\bottomrule
			\end{tabular}
		}
		\caption{Performance if a column is removed from the scheme (detailed results are in Appendix \ref{app:zeroshot}).}
		\vspace{-2mm}
  \label{tab:results_scheme}
	\end{table}

\begin{table}[t!]
		\centering
		\scalebox{0.62}{
			\begin{tabular}{llcccc}
				\toprule
                &\multirow{2}{*}{Method}    & Standard          & \multirow{2}{*}{CoT} & \multirow{2}{*}{Tab-CoT}   \\
                &                           & Prompting         &                    &                      \\
				\midrule
                \multirow{7}*{\scalebox{1}{\hspace*{0.6em}\turnbox{90}{\hspace*{-1.6em}Few-shot} }} &
				SingleEq & 86.8 & {\bf93.1} & 92.1 \\
                &AddSub & {\bf 90.9} & 89.1 & 89.1 \\
                &MultiArith & 44.0 & 96.2 &{\bf96.3}\\
                &GSM8K & 19.7 &{\bf 63.1} & 61.6 \\
                &AQUA & 29.5 & 45.3 & {\bf46.9} \\
                &SVAMP & 69.9 & 76.4 & {\bf82.9} \\
                \cmidrule{2-5}
                &Average & 68.2 & 77.2 & {\bf78.2} \\
				\bottomrule
			\end{tabular}
		}
		\caption{Few-shot results on the arithmetic datasets. }
		\vspace{-2mm}
  \label{tab:results_fewshot}
	\end{table}

\subsection{Importance of Scheme Design}\label{sec:scheme}

To understand the significance of our proposed table scheme design, we evaluate the performance of ``\texttt{|step|subquestion|process|result|}'', along with four variations, each of which is obtained by removing one of the four columns as ablation.
The results in Table \ref{tab:results_scheme} show that each column of ``\texttt{|step|subquestion|process|result|}'' is crucial. 
From the result, we notice that removing the column ``\texttt{step}'' from our scheme results in the most significant performance drop. This implies although the step only contains a number indicating ``which step this is'', it organized the table in sequential order over rows. The column ``\texttt{subquestion}'' is also important. Removing ``subquestion'' from the scheme also shows an average performance drop of 5.4\%. The ``\texttt{subquestion}'' column forms step-by-step instructions vertically, indicating the subquestion under consideration for each step. The ``\texttt{step}'' and ``\texttt{subquestion}'' columns play important roles in maintaining the structure of the table, building vertical connections across rows.

\subsection{Effectiveness of Self-Consistency}

The self-consistency \cite{https://doi.org/10.48550/arxiv.2203.11171} decoding strategy was shown to obtain better results by generating and exploring multiple, diverse reasoning paths.
We also adopt a similar approach here.
In the original self-consistency paper, up to 40 reasoning paths were considered.
We show the feasibility of using only 3 paths in our work.\footnote{The self-consistency decoding method did not show significant improvement when the number of reasoning paths is below 5 in their paper.}
This is conveniently achieved by using 3 different prompts -- we select another two table schemes besides the standard scheme.
One is a highly similar prompt, which we expect to perform similarly well, and the other is less similar, which we expect to yield a worse performance (based on Sec \ref{sec:scheme}).
They are shown in Table \ref{tab:results_vote}.
We then perform majority voting based on the outputs from these 3 prompts.
Interestingly, although a prompt with worse performance is used in the voting process, the overall performance improves.
This shows the benefits of integrating different table schemes for such tasks, which helps improve the overall robustness of the approach.

\subsection{Few-shot Tab-CoT}

Tab-CoT shows impressive reasoning ability under the zero-shot setting.
It can generate a structured output in the form of a table that enables the chain-of-thought reasoning process without few-shot samples. 
Tables are capable chain-of-thought carriers, but can they also serve as good chain-of-thought teachers?
To answer this question, we evaluated Tab-CoT under the few-shot setting.\footnote{We did not compare with least-to-most prompting \cite{https://doi.org/10.48550/arxiv.2205.10625} as it requires task-specific supervision, it only evaluated on GSM8K and provide task-specific prompt for GSM8K in the paper.}

% Few-shot arithmetic reasoning [mix of six, 2-shot comparison]

% While our method with majority voting already has the same average accuracy compared to few-shot standard prompting in arithmetic reasoning, our method shows an impressive performance in the few-shot setting.
% We compared the few-shot abilities with multiple baselines on arithmetic datasets. 

For a fair comparison, we use the same few-shot sample questions described in \citet{https://doi.org/10.48550/arxiv.2201.11903} (listed in Appendix \ref{app:samples}). We use ``\texttt{|step|subquestion|process|result|}'' as the table scheme when representing few-shot samples. The results are reported in Table \ref{tab:results_fewshot}, our method outperformed few-shot CoT by 1\% on average. While the performance difference between Tab-CoT and CoT on other datasets is below 2\%, the performance difference on SVAMP is 6.5\%. The large improvement on SVAMP is likely related to the selection of few-shot samples because \citet{https://doi.org/10.48550/arxiv.2201.11903} select 8 sample questions from SVAMP for all arithmetic reasoning tasks except AQUA\footnote{AQUA has its own few-shot sample questions as they are multiple choice questions. We use the same few-shot samples following \citet{https://doi.org/10.48550/arxiv.2201.11903}.}.
% \footnote{We did not include least-to-most prompting in our experiment because (1) it needs task-specific supervision, (2) it only reports the result in GSM8K and (3) it does not use the same few-shot samples with \citealp{https://doi.org/10.48550/arxiv.2201.11903}, but the selection of few-shot samples will affect the performance.}

\begin{table*}[t!]
		\centering
		\scalebox{0.57}{
            \def\arraystretch{0.7}
			\begin{tabular}{ m{5em}m{3em}m{24em}m{24em}m{10em} } 
				\toprule
				\textbf{Question 1}: &\multicolumn{4}{l}{\textit{Gretchen has some coins. There are 30 more gold coins than silver coins. If she had 70 gold coins, how many coins did Gretchen have in total?} (GT: \textbf{110})} \\
                \midrule
                \multicolumn{1}{l}{\textbf{CoT}:} & \multicolumn{4}{l}{ Let's think step by step. If Gretchen had 70 gold coins, then she would have 30 silver coins (70 - 30 = 40). }\\
                &\multicolumn{4}{l}{ So Gretchen would have a total of \textbf{100} coins (70 + 30 = 100).} \\
				\midrule
                 Model & Step & Subquestion & Process & Result\\
				\midrule
                \multirow{4}*{text}
                &1&How many silver coins does Gretchen have? & There are 30 more gold coins than silver coins. If she had 70 gold coins, how many coins did Gretchen have in total? &100 coins\\
                &2&How many gold coins does Gretchen have? & There are 30 more gold coins than silver coins. If she had 70 gold coins, how many coins did Gretchen have in total? & \textbf{70} gold coins\\
                \midrule
                &&\multicolumn{3}{l}{\textcolor{red}{ \textbf{Error Analysis}: The ``text'' model overwrites the question in the process, thus answering the wrong question.}} \\
				\midrule
				\multirow{3}*{code}
				&1&How many gold coins does Gretchen have?&30 more gold coins than silver coins&70 gold coins\\
                &2&How many silver coins does Gretchen have?&30 more gold coins than silver coins&40 silver coins\\
                &3&How many coins does Gretchen have in total?&70 gold coins + 40 silver coins&\textbf{110} coins\\
                \midrule
                \midrule
                \textbf{Question 2}: &
				\multicolumn{4}{l}{\textit{Adam bought 2 boxes of chocolate candy and 5 boxes of caramel candy. If each box has 4 pieces inside it, how much candy did he have total?} (GT: \textbf{28})} \\
				\midrule
                \multicolumn{1}{l}{\textbf{CoT}:} & \multicolumn{4}{l}{ Adam bought 2 boxes of chocolate candy. That means he has 8 pieces of chocolate candy. Adam bought 5 boxes of caramel candy.  }\\
                &\multicolumn{4}{l}{That means he has 20 pieces of caramel candy. To find out how much candy Adam has in total, } \\
                &\multicolumn{4}{l}{we add 8 pieces of chocolate candy and 20 pieces of caramel candy. Adam has \textbf{28} pieces of candy in total.} \\
				\midrule
                Model & Step & Subquestion & Process & Result\\
				\midrule
                \multirow{8}*{text}
                &1&What is the known information? & Adam bought 2 boxes of chocolate candy and 5 boxes of caramel candy. If each box has 4 pieces inside it, how much candy did he have total? &2 boxes of chocolate candy and 5 boxes of caramel candy, each with 4 pieces inside\\
                &2&What is the unknown information? & How much candy did he have total? & ---\\
                &3&What information is not needed? & --- & ---\\
                &4&What is the equation? &  boxes of chocolate candy and 5 boxes of caramel candy, each with 4 pieces inside & 2(4)+5(4)\\
                &5&What is the answer? & --- & \textbf{24} pieces of candy\\
                \midrule
                &&\multicolumn{3}{l}{\textcolor{red}{\textbf{Error Analysis}: The ``text'' model contains redundant steps in the table and wrong calculation.}} \\
				\midrule
				\multirow{3}*{code}
				&1&How many pieces of chocolate candy did Adam buy?&2 boxes x 4 pieces per box&8 pieces\\
                &2&How many pieces of caramel candy did Adam buy?&5 boxes x 4 pieces per box&20 pieces\\
                &3&How many pieces of candy did Adam buy?&8 pieces + 20 pieces& \textbf{28} pieces\\
                \midrule
                \midrule
                \textbf{Question 3}:&
				\multicolumn{4}{l}{\textit{A pet store had 102 puppies. In one day they sold 21 of them and put the rest into cages with 9 in each cage. How many cages did they use?} (GT: \textbf{9})} \\
				\midrule
                \multicolumn{1}{l}{\textbf{CoT}:} & \multicolumn{4}{l}{ We know that the pet store had 102 puppies and that 21 of them were sold. That means we have 81 puppies left.}\\
                &\multicolumn{4}{l}{Now, if 9 puppies fit in each cage, we can divide 81 by 9 to find out how many cages are needed.81 / 9 = 9. So, the pet store used \textbf{9} cages. } \\
				\midrule
                 Model & Step & Subquestion & Process & Result\\
				\midrule
                \multirow{4}*{text}
                &1&How many cages are needed for 102 puppies? & 102 / 9 = 11.3333 &11\\
                &2&How many cages are needed for 21 puppies? & 21 / 9 = 2.3333 & 2\\
                &3&How many cages are needed in total? & 11 + 2 = 13 & \textbf{13}\\
                \midrule
                &&\multicolumn{3}{l}{\textcolor{red}{\textbf{Error Analysis}: The ``text'' model has a wrong reasoning order, and should add up before dividing.}} \\
				\midrule
				\multirow{3}*{code}
				&1&How many puppies are left?&102 - 21 = 81&81\\
                &2&How many puppies are in each cage?&9&9\\
                &3&How many cages are needed?&81 / 9 = 9& \textbf{9}\\
				\bottomrule
				
			\end{tabular}
		}
		\vspace{-2mm}
		\caption{Case studies (on MultiArith) of the tables generated from ``code-davinci-002''/``text-davinci-002''. The results returned after applying the answer extraction prompts are in \textbf{bold}. Additional case studies are in Appendix \ref{app:casestudy}.}
 	\vspace{-4mm}
\label{tab:results_cs}
	\end{table*}

\subsection{Case Studies}

The main experimental results show that ``code'' under-performs ``text'' with conventional CoT but yields better results in our Tab-CoT.
To understand this better, we conduct case studies to compare their generated tables in Table \ref{tab:results_cs}.

While ``code'' only generated short text snippets or formulas under ``\texttt{process}'', the words generated by ``text'' under the same column tend to form complete sentences whenever possible.
%This proved that the code-based model is more adapted to the structured generation.
As we mentioned earlier, ``code'' is an LLM that is further fine-tuned on code \cite{https://doi.org/10.48550/arxiv.2107.03374}.
This explains why it appears more amenable to the tabular-structured format of the output.
%the preference of generating math equations and concise reasoning is similar to the pattern of code.
In question 1, the model with ``text'' overwrites the generated ``\texttt{subquestion}'' by asking another question.
Thus, the ``\texttt{result}'' fails to answer the ``\texttt{subquestion}'' in the same row.
%Furthermore, not only did ``text'' generate redundant context, but it also generated ambiguous questions.
In question 2, ``text'' generated 5 steps while ``code'' only took 3.
The ``\texttt{subquestion}'' generated by ``text'' is also ambiguous (e.g., ``\texttt{what is the known information?}''). 
In question 3, ``text'' presents a wrong reasoning order.
Overall, ``code'' shows better reasoning ability by demonstrating a more concise and straightforward reasoning process.

\begin{figure*}[h!]
\centering
%  \vspace{4mm}
    \includegraphics[scale=0.38]{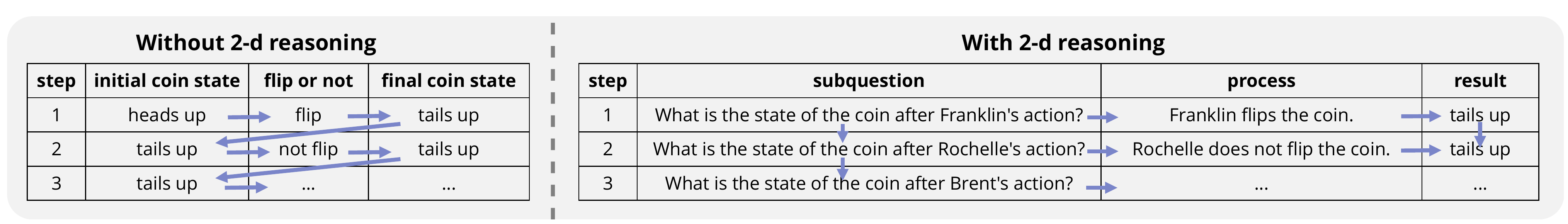}
    \vspace{-3mm}
    \caption{The schemes that disable (left) and enable (right) potential 2-dimensional reasoning.}
    \vspace{-3mm}
    \label{fig:sst_rnn_negate_adj}
\end{figure*}

% general - domain - task
\subsection{Additional Experiments}

We further evaluate our methods on symbolic reasoning and commonsense reasoning tasks. We also conducted some new experiments based on the GPT-3.5 model to understand our approach's effectiveness on such newer models \footnote{GPT-3.5 is released on Mar 2023.}. With such additional experiments, we hope to draw further insights into our approach.

\paragraph{Symbolic Reasoning}

We evaluate Tab-CoT on two symbolic reasoning datasets: Coin Flip (CF)\footnote{An example for Coin Flip: ``\texttt{A coin is heads up. Vinny does not flip the coin. Landon flips the coin. Miguel flips the coin. Caitlyn does not flip the coin. Is the coin still heads up? Note that ``flip'' here means ``reverse''.}''} and Last Letter (LL)\footnote{An example for Last Letter: ``\texttt{Take the Last Letter of each word in ``Vinny Landon Miguel Caitlyn'' and concatenate them.}''}. Unlike the arithmetic reasoning tasks, these tasks focus on some specific problems.
This also opens up the opportunity for us to examine whether the specificity of the table scheme may have an impact on the reasoning process in such tasks.

To this end, we split table schemes into three categories: (1) \textit{general}: the table scheme that can be generally applied to most text questions. (2) \textit{domain-specific}: the table scheme that can be adapted to a specific domain. (3) \textit{task-specific}: the scheme that can only be adopted by a single task.

Our experiments in Table \ref{tab:results_symatic} illustrate that the specificity of the table schemes highly affects the performance of symbolic reasoning tasks.
One may expect the performance to increase as the table scheme becomes more task-specific.
Our task-specific scheme outperformed the zero-shot CoT in both tasks.
However, the increased specificity does not always lead to higher accuracy.
In the Coin Flip task, we noticed that another task-specific scheme ``\texttt{|step|initial coin state|flip or not|next coin state|}'' only achieves an accuracy of 68.0\%.
To understand this, we investigate their reasoning flows in Figure \ref{fig:sst_rnn_negate_adj}. Although the left scheme is more task-specific, it largely disabled the vertical reasoning in the table.
While the right scheme is general, it effectively enables reasoning along both vertical and horizontal directions, leading to significantly better results. \footnote{We further evaluate the general scheme under the one-shot setting, and the results are in Appendix \ref{app:zeroshot}}

\begin{table}[t!]
		\centering
		\scalebox{0.62}{
			\begin{tabular}{ll@{\hskip 0in}c@{\hskip 0in}c@{\hskip 0in}r}
				\toprule
                &Task\ \ \ \  & Cat & Prompt & Result\\
				\midrule
                \multirow{8}*{\scalebox{0.8}{\hspace*{0.6em}\turnbox{90}{\hspace*{-2em}Zero-shot} }} &
				\multirow{4}*{CF}
                &  & {\small \textbf{\texttt{let's think step by step}}} & 91.4 \\
				&& 1 & {\small \textbf{\texttt{|step|subquestion|process|result|}}} & 85.0\\
                % & (1)* & {\small \texttt{|step|subquestion|process|result|}} & 100.0\\
				&& 2 & {\small \textbf{\texttt{|step|initial state|action|next state|}}} & 80.4\\
				% && 3 & {\small \textbf{\texttt{|step|initial coin state|flip or not|next coin state|}}} & 68.0\\
                && 3 & {\small \textbf{\texttt{|step|name|flip or not|result|}}} & \textbf{96.2}\\
				\cmidrule{2-5}
				&\multirow{4}*{LL}
                &  & {\small \textbf{\texttt{let's think step by step}}} & 57.6 \\
				&& 1 & {\small \textbf{\texttt{|step|subquestion|process|result|}}} & 25.2\\
                % & (1)* & {\small \texttt{|step|subquestion|procedure|result|}} & 96.0\\
				&& 2 & {\small \textbf{\texttt{|step|original answer|action|updated answer|}}} & 50.8\\
				&& 3 & {\small \textbf{\texttt{|step|word|last letter|answer|}}} & {\bf 72.8}\\
    %             \midrule
    %             \multirow{2}*{\scalebox{0.8}{\hspace*{0.6em}\turnbox{90}{\hspace*{-1.8em}One-shot} }} &
				% \multirow{1}*{CF}
    %             & 1 & {\small \textbf{\texttt{|step|subquestion|process|result|}}} & 100.0\\
				% \cmidrule{2-5}
				% &\multirow{1}*{LL}
    %             & 1 & {\small \textbf{\texttt{|step|subquestion|process|result|}}} & 96.0\\
                
				\bottomrule
			\end{tabular}
		}
		\caption{Effect of different specificity of schemes. We use Zero-shot CoT with the ``text'' model as our baseline (as Zero-shot CoT works better with ``text'' model).}
        \label{tab:results_symatic}
         % \vspace{-4mm}
	\end{table}

% % main exp table
% \begin{table*}[t!]
% 		\centering
% 		\scalebox{0.6}{
% 			\begin{tabular}{lccccccc|c}
% 				\toprule
%                 Model & Prompt & SingleEq & AddSub & MultiArith & GSM8K & AQUA & SVAMP & Average\\
% 				\midrule
%                 text-davinci-002 & \texttt{|step|question|response|} & 80.1	& 74.9	& 62.2	& 32.2 & 31.9 & 61.6& 60.4\\
% 				code-davinci-002 & \texttt{|step|question|response|} & 77.6& 73.9& 79& 38.1&34.3&63.9& 61.1\\
% 				\midrule
% 				text-davinci-002 & Majority Voting of 3 & 82.7&78.7&75.7&43.8&39.4&63.3&63.9\\
% 				code-davinci-002 & Majority Voting of 3 & 86.4&78.2&85.2&48.2&44.1&66.9&68.2\\
% 				\bottomrule
% 			\end{tabular}
% 		}
% 		\caption{Experiment Result of Tab-CoT.}
% 		\vspace{-2ex}
% 	\end{table*}

\begin{table}[t!]
		\centering
		\scalebox{0.6}{
			\begin{tabular}{llcc|c}
				\toprule
                & Method & CommonsenseQA & StrategyQA & Avg\\
				\midrule
                \multirow{3}*{\scalebox{0.8}{\hspace*{0.6em}\turnbox{90}{\hspace*{-1.5em}Zero-shot} }} &
				Standard Prompting & 69.0 & \textcolor{white}{0}3.3 & 36.2\\
				& CoT & 54.6 & 38.9 & 46.8\\
				&Tab-CoT & 68.4 & 50.4 & 59.4\\
				\cmidrule{1-5}
    %             \multirow{3}*{\scalebox{0.8}{\hspace*{0.6em}\turnbox{90}{\hspace*{-1.5em}Few-shot} }} &
				% Standard Prompting & 82.3 & 67.1 & 74.7\\
				% &CoT & 77.9 & 73.2 & 75.6\\
				% &Tab-CoT & 80.8 & 72.0 & 76.4\\
				% \bottomrule
			\end{tabular}
		}
		\vspace{-2mm}
		\caption{Results on commonsense reasoning.}
		\vspace{-4mm}
  \label{tab:results_commonsense}
	\end{table}

\begin{table}[t!]
		\centering
		\scalebox{0.57}{
			\begin{tabular}{llcc}
				\toprule
                & Method           & CoT & Tab-CoT  \\
				\midrule
                \multirow{7}*{\scalebox{1}{\hspace*{0.8em}\turnbox{90}{\hspace*{-1.6em}Zero-shot} }} &
				SingleEq & 85.6 & \textbf{87.8} \\
                &AddSub & 83.3 & \textbf{85.8 }\\
                &MultiArith & \textbf{90.5} & 89.3 \\
                &GSM8K & 68.7 & \textbf{78.2} \\
                &AQUA & 50.8 & \textbf{51.2} \\
                &SVAMP & 79.0 & \textbf{81.1} \\
                \cmidrule{2-4}
                &Average & 76.3 & \textbf{78.9} \\
				\bottomrule
			\end{tabular}
		}
  \vspace{-2mm}
		\caption{Results with GPT-3.5. }
		\vspace{-4mm}
  \label{tab:results_gpt35}
	\end{table}

\begin{table}[t!]
		\centering
		\scalebox{0.62}{
			\begin{tabular}{llcc}
				\toprule
                &Task & code-cushman-001 & code-davinci-002\\
                \multirow{9}*{\scalebox{1}{\hspace*{0.6em}\turnbox{90}{\hspace*{-1em}Zero-shot} }} &
                &\textcolor{white}{0}(13B)&(175B)\\
				\midrule
				&SingleEq&\textcolor{white}{0}6.3&81.9\\
				&AddSub&\textcolor{white}{0}6.3&70.9\\
				&MultiArith&\textcolor{white}{0}2.0&81.2\\
				&GSM8K&\textcolor{white}{0}0.9&44.4\\
				&AQUA&16.9&37.0\\
				&SVAMP&\textcolor{white}{0}5.0&60.5\\
                \cmidrule{2-4}
                &\textbf{Average}&\textcolor{white}{0}6.2&62.6\\
				\bottomrule
			\end{tabular}
		}
		\caption{A comparison between the different sizes of ``code'', ``Average'' is the average score across six datasets.}
  \vspace{-4mm}
  \label{tab:results_modelsize}
	\end{table}

 \begin{table*}[t!]
        % \vspace{4mm}
		\centering
		\scalebox{0.57}{
			\begin{tabular}{llcccccc|c}
				\toprule
                & Scheme & SingleEq & AddSub & MultiArith & GSM8K & AQUA & SVAMP & Average\\
                \midrule
                \multirow{5}*{\hspace*{0.6em}\turnbox{90}{\hspace*{-2em}Zero-shot} } &
                Standard Prompting & 46.3 &	51.4&\textcolor{white}{0}7.2&\textcolor{white}{0}4.1&23.6&29.5 & 27.0 \\
				\cmidrule{2-9}
				&{\texttt{|subquestion|process|result|}} &69.9	&51.9	&84.0	&40.1	&35	&44.7		&54.3\\
				&{\texttt{|step|process|result|}} &77.0	&55.7	&84.2	&41.5	&37.8	&46.9		&57.2\\
				&{\texttt{|step|subquestion|result|}} &76.0	&77.9	&76.8	&40.1	&36.2	&60.6		&61.3\\
				&{\texttt{|step|subquestion|process|}} &78.0	&75.9	&76.3	&39.7	&34.3	&60.9		&60.9\\
				\cmidrule{2-9}
				&{\texttt{|step|subquestion|process|result|}} & 81.9&70.9&81.2&44.4&37.0&60.5  & {\bf 62.6}\\
				\bottomrule
			\end{tabular}
		}
		\caption{Performance if a column is removed from the scheme.}
    \vspace{-4mm}
  \label{tab:results_scheme_appendix}
	\end{table*}

\paragraph{Commonsense Reasoning}

As another set of additional experiments, we further evaluate our method on commonsense reasoning, including CommonsenseQA \cite{talmor-etal-2019-commonsenseqa} and StrategyQA \cite{geva-etal-2021-aristotle}. The results are in Table \ref{tab:results_commonsense}.
Tab-CoT obtained the highest average accuracy. However, the results of our method did not show significantly improved performance compared with Standard Prompting in a few-shot setting\footnote{The few-shot results are in Appendix \ref{app:fewshot}.}. These results imply that commonsense reasoning tasks do not have a fixed answering pattern. Therefore, providing chain-of-thought samples is not enough to make up for the lack of commonsense knowledge. For a fair comparison, we use the same few-shot questions listed in \cite{https://doi.org/10.48550/arxiv.2201.11903}. 

\paragraph{Results on GPT-3.5}

We test our method on the recent model ``GPT-3.5-turbo-0301'' in Table \ref{tab:results_gpt35}\footnote{Our experiment is conducted in May 2023. The ``GPT-3.5-turbo-0301'' may be updated in the future.}. We found that our method is applicable to GPT-3.5, and achieves better performance compared to conventional Zero-shot CoT. Another interesting observation is when prompting the GPT-3.5 model with ``Let's think step by step'', a large number of the generated texts already contain a table in their CoT process. \footnote{Based on our observations, those tables generated in conventional Zero-shot CoT under GPT 3.5 can be different from those generated with our method. They appear to be mostly used to organize information related to the question but do not appear to be used for presenting reasoning steps.}

\subsection{Ablation Studies}

\paragraph{Model Sizes}

\citet{llmrzr} evaluated the family of GPT-3 models of four different sizes: 2.7B, 6.7B, 13B, and 175B parameters. The results show that only the largest model (``text-davinci-002'') shows the chain-of-thought reasoning ability.

We compare the performance of the smaller model ``code-cushman-001'' (13B) with ``code-davinci-002'' (175B). Similar to zero-shot CoT, smaller models do not show the ability to conduct chain-of-thought reasoning.
The performance of ``code-cushman-001'' cannot reach 10\%, except AQUA (a multiple choice dataset with 5 choices for each question).
The experimental results are reported in Table \ref{tab:results_modelsize}.

\paragraph{Structure-Promoting Scheme}

As mentioned in Table \ref{tab:results_scheme}, we compare the performance when we remove any column from ``\texttt{|step|subquestion|process|result|}''. The detailed experimental results are reported in Table \ref{tab:results_scheme_appendix}. Results suggest that each column of our proposed scheme is important because removing any column will lead to a drop in performance.

\section{Discussion}

Our experimental results confirmed the effectiveness of our proposed tabular chain-of-thought method under both zero-shot and few-shot settings.
We summarize several advantages of our method compared to conventional chain-of-thought methods and list them below.

Tab-CoT generates a table illustrating the reasoning process, which is more {\em organized}.
This nature of the generated text, as can be seen from Table \ref{tab:results_cs}, makes the reasoning process much easier.

Additionally, from Figure \ref{fig:sst_rnn_negate_adj}, we conclude that Tab-CoT encourages a more {\em structured} reasoning process to be explicitly modelled.
As a 2-dimensional data structure, tables enable both horizontal reasoning along rows and vertical reasoning along columns.

Practically, table schemes are also {\em easy to craft}. Designing a specific table generation prompt typically involves deciding concise header names without concerning grammar. It is thus less cumbersome than choosing a natural language prompt from a diverse set of candidates.

Overall, we argue that under current state-of-the-art LLMs, table schemes are {\em natural prompts} that are well suited for zero-shot learning.
%Other few-shot prompting methods may be difficult to apply in a zero-shot setting, as their pattern is implicit and needs to be learned with few-shot samples. However, a table is included in the pre-trained data and has an explicit structure, which can be adapted by LLMs without few-shot samples more naturally.

\section{Conclusion}
% table as prompt
In this paper, we propose Tab-CoT, a novel prompting framework that performs effective zero-shot reasoning by generating a table.

Tab-CoT shows competitive results on arithmetic reasoning tasks under both zero-shot and few-shot settings.
We further conducted comprehensive experiments across different reasoning tasks under different settings.
Our comprehensive experiments revealed some specific benefits of our method and identify the optimal way to use it.
We hope that, through our work, we can sparkle new ideas and provide some inspiration to our community.

In the future, we would like to explore methods to automate the scheme selection process, using the generated schemes to meet task-specific requirements. Future work also includes integrating external calculators \cite{https://doi.org/10.48550/arxiv.2211.10435}, or task-specific supervision \cite{https://doi.org/10.48550/arxiv.2205.10625} into the learning process, under both zero-shot and few-shot settings.

Our Tab-CoT also provides a straightforward decomposition of the intermediate thought process. This highly structured chain of thought produced by our approach may help people to observe and interpret how large language models decompose complex problems. We believe our proposed method can help reveal the underlying mechanisms associated with the emergence of certain complex behaviours associated with large language models.

% For future work, we can automate the scheme selection process, using the generated schemes for task-specific requirements. Future work also includes integrating external calculator \cite{https://doi.org/10.48550/arxiv.2211.10435}, or task-specific supervision \cite{https://doi.org/10.48550/arxiv.2205.10625} into the learning process, under both zero-shot and few-shot settings.

% Our Tab-CoT also provides a straightforward decomposition of chain-of-thought. This highly structured chain of thought may help people to observe and interpret how large language models decompose complex problems. The table parse such chain-of-thought well, as each step takes a row, we argue that further development like a search algorithm can be easily taken place.
%These add-ons may also be adapted in Tab-CoT for better performance.

\section*{Limitations}
% ok
We identify a few limitations of this work. First, our approach is applicable to language models pre-trained with tables, which may not always be included in all language models (especially small ones).
Second, our approach's limited improvement in commonsense reasoning tasks suggests that its effectiveness may depend on the specific task and the level of structured reasoning required.

\section*{Acknowledgement}

We would like to thank the anonymous reviewers, our meta-reviewer, and senior area chairs for their constructive comments and support on our work.
This research/project is supported by the National Research Foundation Singapore and DSO National Laboratories under the AI Singapore Program (AISG Award No: AISG2-RP-2020-016), and the Ministry of Education, Singapore, under its Tier 3 Programme (The Award No.: MOET32020-0004).

\bibliography{custom}
\bibliographystyle{acl_natbib}

\appendix

\section{One-shot Reasoning on Symbolic Reasoning}
\label{app:zeroshot}
\label{app:symbolic}
We evaluate our method on Coin Flip (Table \ref{tab:few_shot_coin}) and Last Letter (Table \ref{tab:few_shot_last}) under the one-shot setting. As shown in Table \ref{tab:results_symatic_extra}, by adding one few-shot sample, LLMs can gain a significant performance boost in both tasks with general scheme ``\texttt{|step|subquestion|process|result|}''.

\section{Additional Few-shot Results}
\label{app:fewshot}

We evaluate our method on commonsense reasoning tasks under a few-shot setting. Our model performs slightly better in terms of average accuracy. The results are reported in Table \ref{tab:results_commonsense_extra}.

\section{Additional Case Studies}
\label{app:casestudy}

We show some errors our method made in arithmetic reasoning tasks through further case studies. The results are reported in Table \ref{tab:case_study_extra} and \ref{tab:case_study_extra2}.

\section{Few-Shot Samples}
\label{app:samples}

We list our few-shot samples for all arithmetic reasoning (Table \ref{tab:few_shot_math} and Table \ref{tab:few_shot_aqua}), CommonsenseQA (Table \ref{tab:few_shot_csqa}) and StrategyQA (Table \ref{tab:few_shot_strat}).
We use the same few-shot sample questions from \citet{https://doi.org/10.48550/arxiv.2201.11903} 
 
\begin{table*}[t!]
		\centering
		\scalebox{0.6}{
			\begin{tabular}{llcc|c}
				\toprule
                & Method & CommonsenseQA & StrategyQA & Avg\\
				\midrule
    %             \multirow{3}*{\scalebox{0.8}{\hspace*{0.6em}\turnbox{90}{\hspace*{-1.5em}Zero-shot} }} &
				% Standard Prompting & 69.0 & \textcolor{white}{0}3.3 & 36.2\\
				% & CoT & 54.6 & 38.9 & 46.8\\
				% &Tab-CoT & 68.4 & 50.4 & 59.4\\
				% \cmidrule{1-5}
                \multirow{3}*{\scalebox{0.8}{\hspace*{0.6em}\turnbox{90}{\hspace*{-1.5em}Few-shot} }} &
				Standard Prompting & 82.3 & 67.1 & 74.7\\
				&CoT & 77.9 & 73.2 & 75.6\\
				&Tab-CoT & 80.8 & 72.0 & 76.4\\
				\bottomrule
			\end{tabular}
		}
		\caption{Few-shot results on commonsense reasoning.}
  \label{tab:results_commonsense_extra}
	\end{table*}

\begin{table*}[t]
		\centering
		\scalebox{0.62}{
			\begin{tabular}{llccr}
				\toprule
                &Task\ \ \ \  & Prompt & Result\\
				\midrule
             \multirow{2}*{\scalebox{0.6}{\hspace*{0.6em}\turnbox{90}{\hspace*{-1.5em}Zero-shot} }} &
                \multirow{1}*{CF}
    %             &  & {\small \textbf{\texttt{let's think step by step}}} & 91.4 \\
				 & {\small \textbf{\texttt{|step|subquestion|process|result|}}} & 85.0\\
    %             % & (1)* & {\small \texttt{|step|subquestion|process|result|}} & 100.0\\
				% && 2 & {\small \textbf{\texttt{|step|initial state|action|next state|}}} & 80.4\\
				% % && 3 & {\small \textbf{\texttt{|step|initial coin state|flip or not|next coin state|}}} & 68.0\\
    %             && 3 & {\small \textbf{\texttt{|step|name|flip or not|result|}}} & \textbf{96.2}\\
				% \cmidrule{2-5}
				 &\multirow{1}*{LL}
    %             &  & {\small \textbf{\texttt{let's think step by step}}} & 57.6 \\
				& {\small \textbf{\texttt{|step|subquestion|process|result|}}} & 25.2\\
    %             % & (1)* & {\small \texttt{|step|subquestion|procedure|result|}} & 96.0\\
				% && 2 & {\small \textbf{\texttt{|step|original answer|action|updated answer|}}} & 50.8\\
				% && 3 & {\small \textbf{\texttt{|step|word|last letter|answer|}}} & {\bf 72.8}\\
    %             \midrule
                 \cmidrule{2-4}
                \multirow{2}*{\scalebox{0.6}{\hspace*{0.6em}\turnbox{90}{\hspace*{-1.5em}One-shot} }} &
				\multirow{1}*{CF}
                & {\small \textbf{\texttt{|step|subquestion|process|result|}}} & 100.0\\
				&\multirow{1}*{LL}
                & {\small \textbf{\texttt{|step|subquestion|process|result|}}} & 96.0\\
                
				\bottomrule
			\end{tabular}
		}
		\caption{One-shot results on symbolic reasoning.}
        \label{tab:results_symatic_extra}
         % \vspace{-4mm}
	\end{table*}
 
    \begin{table*}[h]
		\centering
		\scalebox{0.6}{
			\begin{tabular}{ m{64em} } 
				\toprule
                Case Study\\
				\midrule

\textbf{Tom bought a skateboard for \$ 9.46 , and spent \$ 9.56 on marbles . Tom also spent \$ 14.50 on shorts . In total , how much did Tom spend on toys ?}\\
\\
|step|subproblem|process|result|\\
|---|---|---|---|\\
|1|How much did Tom spend on a skateboard?|\$9.46|\$9.46|\\
|2|How much did Tom spend on marbles?|\$9.56|\$9.56|\\
|3|How much did Tom spend on shorts?|\$14.50|\$14.50|\\
|4|How much did Tom spend on toys?|\$9.46 + \$9.56 + \$14.50|\$33.52|
Therefore, the answer (arabic numerals) is \$33.52. \textcolor{red}{(GT: 19.02)}\\
				\midrule
				\textcolor{red}{\textbf{Error Analysis: Misleading by redundant information. Regard shorts as toys.}}\\
				\midrule

\textbf{Tom found 15 seashells and Fred found 43 seashells on the beach . When they cleaned them , they discovered that 29 were cracked . How many seashells did they find together ?}\\
\\
|step|subproblem|process|result|\\
|---|---|---|---|\\
|1|How many seashells did Tom find?|15|15|\\
|2|How many seashells did Fred find?|43|43|\\
|3|How many seashells were cracked?|29|29|\\
|4|How many seashells did they find together?|15 + 43|58|\\
|5|How many seashells did they keep?|58 - 29|29|\\
Therefore, the answer (arabic numerals) is 29. \textcolor{red}{ (GT: 58)}\\ 
				\midrule
				\textcolor{red}{\textbf{Error Analysis: Misleading by redundant information. Cracked seashells are still seashells.}}\\
				\midrule

\textbf{The grasshopper, the frog and the mouse had a jumping contest. The grasshopper jumped 19 inches. The grasshopper jumped 4 inches farther than the frog and the mouse jumped 44 inches lesser than the frog. How far did the frog jump?}\\
\\
|step|subproblem|process|result|\\
|---|---|---|---|\\
|1|How far did the grasshopper jump?|19|19|\\
|2|How far did the grasshopper jump farther than the frog?|4|4|\\
|3|How far did the mouse jump lesser than the frog?|44|44|\\
|4|How far did the frog jump?|19 - 4|15|\\
|5|How far did the frog jump?|19 + 44|63|\\
|6|How far did the frog jump?|63 - 15|48|\\
Therefore, the answer (arabic numerals) is 48. \textcolor{red}{ (GT: 15)}\\ 
				\midrule
				\textcolor{red}{\textbf{Error Analysis: Misleading by redundant information. The mouse does not matter.}}\\
				\bottomrule
			\end{tabular}
		}
		\caption{Few-shot case studies in arithmetic reasoning datasets.}
		\label{tab:case_study_extra}
	\end{table*}

    \begin{table*}[h]
		\centering
		\scalebox{0.6}{
			\begin{tabular}{ m{64em} } 
				\toprule
                Case Study\\
				\midrule
\textbf{A coin is heads up. Whitney flips the coin. Erika does not flip the coin. Tj does not flip the coin. Benito flips the coin. Is the coin still heads up? Note that ``flip'' here means ``reverse''.}\\
\\
|step|initial coin state|flip or not|next coin state|\\
|---|---|---|---|\\
|1|heads up|no flip|heads up|\\
|2|heads up|no flip|heads up|\\
|3|heads up|flip|tails up|\\
Therefore, the answer (Yes or No) is No. \textcolor{red}{ (GT: Yes)}\\
				\midrule
				\textcolor{red}{\textbf{Error Analysis: The Table failed to simulate all 4 steps and was early-stopped.}}\\
				\bottomrule
			\end{tabular}
		}
		\caption{Case studies in symbolic reasoning datasets.}
		\label{tab:case_study_symbolic}
	\end{table*}

    \begin{table*}[h]
		\centering
		\scalebox{0.6}{
			\begin{tabular}{ m{64em} } 
				\toprule
                Case Study\\
				\midrule

\textbf{A bakery produces 60 loaves of bread each day. Two-thirds of the loaves are sold in the morning and half of what is left is sold equally in the afternoon and evening. How many loaves of bread are sold in the afternoon?}\\
\\
|step|subquestion|process|result|\\
|---|---|---|---|\\
|1|How many loaves of bread are produced each day?|60 loaves|60 loaves|\\
|2|How many loaves of bread are sold in the morning?|2/3 * 60 loaves = 40 loaves|40 loaves|\\
|3|How many loaves of bread are left?|60 loaves - 40 loaves = 20 loaves|20 loaves|\\
|4|How many loaves of bread are sold in the afternoon and evening?|1/2 * 20 loaves = 10 loaves|10 loaves|\\
|5|How many loaves of bread are sold in the afternoon?|10 loaves / 2 = 5 loaves|5 loaves|
Therefore, the answer (arabic numerals) is 5.\textcolor{red}{ (GT: 10)}\\
				\midrule
				\textcolor{red}{\textbf{Error Analysis: Wrong subquestion in step 4. This result in /2 in step 5.}}\\
				\midrule

\textbf{John hires a driving service to get him to work each day.  His work is 30 miles away and he has to go there and back each day.  He goes to work 5 days a week for 50 weeks a year.  He gets charged $2 per mile driven and he also gives his driver a $150 bonus per month.  How much does he pay a year for driving?}\\
\\
|step|subquestion|process|result|\\
|---|---|---|---|\\
|1|How many miles does he drive each day?|30 miles|30 miles|\\
|2|How many days a week does he go to work?|5 days|5 days|\\
|3|How many weeks a year does he go to work?|50 weeks|50 weeks|\\
|4|How many miles does he drive a year?|30 miles * 5 days * 50 weeks = 7,500 miles|7,500 miles|\\
|5|How much does he pay per mile?|\$2|\$2|\\
|6|How much does he pay for driving a year?|\$2 * 7,500 miles = \$15,000|\$15,000|\\
|7|How much does he pay for the bonus?|\$150 * 12 months = \$1,800|\$1,800|\\
|8|How much does he pay a year for driving?|\$15,000 + \$1,800 = \$16,800|\$16,800|\\
Therefore, the answer (arabic numerals) is \$16,800. \textcolor{red}{ (GT: 31800)}\\
				\midrule
				\textcolor{red}{\textbf{Error Analysis: Forget that John need to go there and back, so it is a two-way-trip.}}\\
				\midrule

\textbf{There are four birds at the Caboose. Sally Two is three years older than Granny Red. Granny Red is two times as old as Sally Four. If Sally Four is the same age as Sally Thirtytwo, and Sally Thirtytwo is 8 years old, what's the total age of the four birds?}\\
\\
|step|subquestion|process|result|\\
|---|---|---|---|\\
|1|How old is Sally Four?|8 years|8 years|\\
|2|How old is Sally Thirtytwo?|8 years|8 years|\\
|3|How old is Granny Red?|8 years / 2 = 4 years|4 years|\\
|4|How old is Sally Two?|4 years + 3 years = 7 years|7 years|\\
|5|What's the total age of the four birds?|8 years + 8 years + 4 years + 7 years = 27 years|27 years|\\
Therefore, the answer (arabic numerals) is 27.\textcolor{red}{ (GT: 27)}\\
				\midrule
				\textcolor{red}{\textbf{Error Analysis: Step 3, Granny is 2 times older, so it is *2.}}\\
				\midrule

\textbf{Sam 's dog had puppies and 8 had spots . He gave 2 to his friends . He now has 6 puppies . How many puppies did he have to start with ?}\\
\\
|step|subproblem|process|result|\\
|---|---|---|---|\\
|1|How many puppies did Sam have to start with?|8|8|\\
|2|How many puppies did he give to his friends?|2|2|\\
|3|How many puppies does he have now?|6|6|\\
|4|How many puppies did he have to start with?|8 - 2|6|\\
Therefore, the answer (arabic numerals) is 6.\textcolor{red}{ (GT: 8)}\\
				\midrule
				\textcolor{red}{\textbf{Error Analysis: The question ask for existing value, but the table still do the calculation.}}\\
				\midrule

\textbf{The school is planning a field trip. The school has 87 classrooms. There are 58 students in the school with each classroom having the same number of students. If there are 2 seats on each school bus. How many buses are needed to take the trip?}\\
\\
|step|subproblem|process|result|\\
|---|---|---|---|\\
|1|How many classrooms are in the school?|87|87|\\
|2|How many students are in the school?|58|58|\\
|3|How many students are in each classroom?|58 / 87|0.67|\\
|4|How many seats are on each bus?|2|2|\\
|5|How many students can fit on each bus?|2 * 0.67|1.34|\\
|6|How many buses are needed to take the trip?|58 / 1.34|43|\\
Therefore, the answer (arabic numerals) is 43.\textcolor{red}{ (GT: 29)}\\
				\midrule
				\textcolor{red}{\textbf{Error Analysis: Redundant steps by redundant information, 58/2=29 .}}\\
				\bottomrule
			\end{tabular}
		}
		\caption{Additional few-shot Case studies in arithmetic reasoning datasets.}
		\label{tab:case_study_extra2}
	\end{table*}

\begin{table*}[h]
		\centering
		\scalebox{0.6}{
			\begin{tabular}{ m{64em} } 
				\toprule
                Few-shot Samples For 5 Arithmetic Reasoning Tasks\\
				\midrule
				{\bf There are 15 trees in the grove. Grove workers will plant trees in the grove today. After they are done, there will be 21 trees. How many trees did the grove workers plant today?}\\
                \\
                |step|subquestion|process|result|\\
                |---|---|---|---|\\
                |1|How many trees are in the grove?|15|15|\\
                |2|How many trees will be in the grove after the workers are done?|21|21|\\
                |3|How many trees did the workers plant?|21 - 15|6|\\
                Therefore, the answer (arabic numerals) is 6.\\
                \\
               {\bf If there are 3 cars in the parking lot and 2 more cars arrive, how many cars are in the parking lot?}\\
                \\
                |step|subquestion|process|result|\\
                |---|---|---|---|\\
                |1|How many cars are in the parking lot?|3|3|\\
                |2|How many cars arrive?|2|2|\\
                |3|How many cars are in the parking lot?|3 + 2|5|\\
                Therefore, the answer (arabic numerals) is 5.\\
                \\
                {\bf Leah had 32 chocolates and her sister had 42. If they ate 35, how many pieces do they have left in total?}\\
                \\
                |step|subquestion|process|result|\\
                |---|---|---|---|\\
                |1|How many chocolates did Leah have?|32|32|\\
                |2|How many chocolates did her sister have?|42|42|\\
                |3|How many chocolates did they eat?|35|35|\\
                |4|How many chocolates do they have left?|32 + 42 - 35|39|\\
                Therefore, the answer (arabic numerals) is 39.\\
                \\
                {\bf Jason had 20 lollipops. He gave Denny some lollipops. Now Jason has 12 lollipops. How many lollipops did Jason give to Denny?}\\
                \\
                |step|subquestion|process|result|\\
                |---|---|---|---|\\
                |1|How many lollipops did Jason have?|20|20|\\
                |2|How many lollipops does Jason have now?|12|12|\\
                |3|How many lollipops did Jason give to Denny?|20 - 12|8|\\
                Therefore, the answer (arabic numerals) is 8.\\
                \\
                {\bf Shawn has five toys. For Christmas, he got two toys each from his mom and dad. How many toys does he have now?}\\
                \\
                |step|subquestion|process|result|\\
                |---|---|---|---|\\
                |1|How many toys does Shawn have?|5|5|\\
                |2|How many toys did he get from his mom?|2|2|\\
                |3|How many toys did he get from his dad?|2|2|\\
                |4|How many toys does he have now?|5 + 2 + 2|9|\\
                Therefore, the answer (arabic numerals) is 9.\\
                \\
                {\bf There were nine computers in the server room. Five more computers were installed each day, from monday to thursday. How many computers are now in the server room?}\\
                \\
                |step|subquestion|process|result|\\
                |---|---|---|---|\\
                |1|How many computers were in the server room?|9|9|\\
                |2|How many computers were installed each day?|5|5|\\
                |3|How many computers were installed from monday to thursday?|5 * 4|20|\\
                |4|How many computers are now in the server room?|9 + 20|29|\\
                Therefore, the answer (arabic numerals) is 29.\\
                \\
                {\bf Michael had 58 golf balls. On tuesday, he lost 23 golf balls. On wednesday, he lost 2 more. How many golf balls did he have at the end of wednesday?}\\
                \\
                |step|subquestion|process|result|\\
                |---|---|---|---|\\
                |1|How many golf balls did Michael have?|58|58|\\
                |2|How many golf balls did he lose on tuesday?|23|23|\\
                |3|How many golf balls did he lose on wednesday?|2|2|\\
                |4|How many golf balls did he have at the end of wednesday?|58 - 23 - 2|33|\\
                Therefore, the answer (arabic numerals) is 33.\\
                \\
                {\bf Olivia has \$23. She bought five bagels for \$3 each. How much money does she have left?}\\
                \\
                |step|subquestion|process|result|\\
                |---|---|---|---|\\
                |1|How much money does Olivia have?|\$23|\$23|\\
                |2|How much does each bagel cost?|\$3|\$3|\\
                |3|How many bagels did she buy?|5|5|\\
                |4|How much money did she spend on bagels?|\$3 * 5|\$15|\\
                |5|How much money does she have left?|\$23 - \$15|\$8|\\
                Therefore, the answer (arabic numerals) is \$8.\\
				\midrule
				\bottomrule
			\end{tabular}
		}
		\caption{Few-shot samples for 5 arithmetic reasoning tasks, including SingleEq, AddSub, MultiArith, GSM8K, and SVAMP}
		\label{tab:few_shot_math}
	\end{table*}
	
	\begin{table*}[t]
		\centering
		\scalebox{0.6}{
			\begin{tabular}{ m{64em} } 
				\toprule
                Few-shot Samples For AQUA\\
				\midrule
				\midrule
				{\bf John found that the average of 15 numbers is 40. If 10 is added to each number then the mean of the numbers is? Answer Choices: (A) 50 (B) 45 (C) 65 (D) 78 (E) 64}\\
                \\
                |step|subquestion|process|result|\\
                |---|---|---|---|\\
                |1|How much did the new mean change?|If 10 is added to each number, then the mean of the numbers also increases by 10.|10|\\
                |2|What is the new mean?|So the new mean would be 40 + 10 = 50.|50|\\
                Therefore, among A through E, the answer is A.\\
                \\
                {\bf If a / b = 3/4 and 8a + 5b = 22, then find the value of a. Answer Choices: (A) 1/2 (B) 3/2 (C) 5/2 (D) 4/2 (E) 7/2}\\
                \\
                |step|subquestion|process|result|\\
                |---|---|---|---|\\
                |1|What equation we have have if we substitute b with a?| a / b = 3/4, then b = 4a / 3. So 8a + 5(4a / 3) = 22.|8a + 5(4a / 3) = 22|\\
                |2|What is the value of a?|8a + 5(4a / 3) = 22 simplifies to 8a + 20a / 3 = 22, which means 44a / 3 = 22. So a is equal to 3/2.|2/3|\\
                Therefore, among A through E, the answer is B.\\
                \\
               {\bf  A person is traveling at 20 km/hr and reached his destiny in 2.5 hr then find the distance? Answer Choices: (A) 53 km (B) 55 km (C) 52 km (D) 60 km (E) 50 km}\\
                \\
                |step|subquestion|process|result|\\
                |---|---|---|---|\\
                |1|What is the distance this person traveling?|The distance that the person traveled would have been 20 km/hr * 2.5 hrs = 50 km.|50km|\\
                Therefore, among A through E, the answer is E.\\
                \\
                {\bf How many keystrokes are needed to type the numbers from 1 to 500? Answer Choices: (A) 1156 (B) 1392 (C) 1480 (D) 1562 (E) 1788}\\
                \\
                |step|subquestion|process|result|\\
                |---|---|---|---|\\
                |1|How many one-digit numbers are there?|There are 9 one-digit numbers from 1 to 9.|9|\\
                |2|How many two-digit numbers are there?|There are 90 two-digit numbers from 10 to 99.|90|\\
                |3|How many three-digit numbers are there?|There are 401 three-digit numbers from 100 to 500.|401|\\
                |4|How many keystrokes are needed to type the number from 1 to 500?|9 + 90(2) + 401(3) = 1392.|1392|\\
                Therefore, among A through E, the answer is B.\\
				\bottomrule
			\end{tabular}
		}
		\caption{Few-shot samples for AQUA}
		\label{tab:few_shot_aqua}
	\end{table*}
	
		\begin{table*}[t]
		\centering
		\scalebox{0.6}{
			\begin{tabular}{ m{64em} } 
				\toprule
                Few-shot Samples For CommonsenseQA\\
				\midrule
				\midrule
				{\bf What do people use to absorb extra ink from a fountain pen? Answer Choices: (A) shirt pocket (B) calligrapher’s hand (C) inkwell (D) desk drawer (E) blotter}\\
                \\
                |step|subquestion|process|result|\\
                |---|---|---|---|\\
                |1|What can we know of answer?|The answer must be an item that can absorb ink.|(E)|
                Therefore, Among A through E, the answer is E.\\
                \\
                {\bf What home entertainment equipment requires cable? Answer Choices: (A) radio shack (B) substation (C) television (D) cabinet}\\
                \\
                |step|subquestion|process|result|\\
                |---|---|---|---|\\
                |1|What can we know of answer?|The answer must require cable.|(C)|\\
                Therefore, Among A through E, the answer is C.\\
                \\
                {\bf The fox walked of city into the forest, what was it looking for? Answer Choices: (A) pretty flowers (B) hen house (C) natural habitat (D) storybook}\\
                \\
                |step|subquestion|process|result|\\
                |---|---|---|---|\\
                |1|What can we know of answer?|The answer must be something in the forest.|(B)|\\
                Therefore, Among A through E, the answer is B.\\
                \\
                {\bf Sammy wanted to go to where the people were. Where might he go? Answer Choices: (A) populated areas (B) race track (C) desert (D) apartment (E) roadblock}\\
                \\
                |step|subquestion|process|result|\\
                |---|---|---|---|\\
                |1|What can we know of answer?|The answer must be a place with a lot of people.|(A)|\\
                Therefore, Among A through E, the answer is A.\\
                \\
                {\bf Where do you put your grapes just before checking out? Answer Choices: (A) mouth (B) grocery cart (C)super market (D) fruit basket (E) fruit market}\\
                \\
                |step|subquestion|process|result|\\
                |---|---|---|---|\\
                |1|What can we know of answer?|The answer should be the place where grocery items are placed before checking out.|(B)|\\
                Therefore, Among A through E, the answer is B.\\
                \\
                {\bf Google Maps and other highway and street GPS services have replaced what? Answer Choices: (A) united states (B) mexico (C) countryside (D) atlas}\\
                \\
                |step|subquestion|process|result|\\
                |---|---|---|---|\\
                |1|What can we know of answer?|The answer must be something that used to do what Google Maps and GPS services do, which is to give directions.|(D)|\\
                Therefore, Among A through E, the answer is D.\\
                \\
                {\bf Before getting a divorce, what did the wife feel who was doing all the work? Answer Choices: (A) harder (B) anguish (C) bitterness (D) tears (E) sadness}\\
                \\
                |step|subquestion|process|result|\\
                |---|---|---|---|\\
                |1|What can we know of answer?|The answer should be the feeling of someone getting divorced who was doing all the work.|(C)|\\
                Therefore, Among A through E, the answer is C.\\

				\bottomrule
			\end{tabular}
		}
		\caption{Few-shot samples for CommonsenseQA}
		\label{tab:few_shot_csqa}
	\end{table*}
	
		\begin{table*}[t]
		\centering
		\scalebox{0.6}{
			\begin{tabular}{ m{64em} } 
				\toprule
                Few-shot Samples For StrategyQA\\
				\midrule
				\midrule
				{\bf Do hamsters provide food for any animals?}\\
                \\
                |step|subquestion|process|result|\\
                |---|---|---|---|\\
                |1|What is the evidence?|Hamsters are prey animals. Prey are food for predators.|yes|\\
                Therefore, the answer (yes or no) is yes.\\
                \\
                {\bf Could Brooke Shields succeed at University of Pennsylvania?}\\
                \\
                |step|subquestion|process|result|\\
                |---|---|---|---|\\
                |1|What is the evidence?|Brooke Shields went to Princeton University. Princeton University is about as academically rigorous as the University of Pennsylvania. Thus, Brooke Shields could also succeed at the University of Pennsylvania.|yes|\\
                Therefore, the answer (yes or no) is yes.\\
                \\
                {\bf Yes or no: Hydrogen’s atomic number squared exceeds number of Spice Girls?}\\
                \\
                |step|subquestion|process|result|\\
                |---|---|---|---|\\
                |1|What is the evidence?|Hydrogen has an atomic number of 1. 1 squared is 1. There are 5 Spice Girls. Thus, Hydrogen’s atomic number squared is less than 5.|no|\\
                Therefore, the answer (yes or no) is no.\\
                \\
                {\bf Yes or no: Is it common to see frost during some college commencements?}\\
                \\
                |step|subquestion|process|result|\\
                |---|---|---|---|\\
                |1|What is the evidence?|College commencement ceremonies can happen in December, May, and June. December is in the winter, so there can be frost. Thus, there could be frost at some commencements.|yes|\\
                Therefore, the answer (yes or no) is yes.\\
                \\
                {\bf Yes or no: Could a llama birth twice during War in Vietnam (1945-46)?}\\
                \\
                |step|subquestion|process|result|\\
                |---|---|---|---|\\
                |1|What is the evidence?|The War in Vietnam was 6 months. The gestation period for a llama is 11 months, which is more than 6 months. Thus, a llama could not give birth twice during the War in Vietnam.|no|\\
                Therefore, the answer (yes or no) is no.\\
                \\
                {\bf Yes or no: Would a pear sink in water?}\\
                \\
                |step|subquestion|process|result|\\
                |---|---|---|---|\\
                |1|The density of a pear is about 0.6g/cm3, which is less than water. Objects less dense than water float. Thus, a pear would float.|no|\\
                Therefore, the answer (yes or no) is no.\\

				\bottomrule
			\end{tabular}
		}
		\caption{Few-shot samples for StrategyQA}
		\label{tab:few_shot_strat}
	\end{table*}

		\begin{table*}[t]
		\centering
		\scalebox{0.6}{
			\begin{tabular}{ m{64em} } 
				\toprule
                One-shot Sample Used on Coin Flip\\
				\midrule
				\midrule
                {\bf A coin is heads up. Dorian flips the coin. Mayra flips the coin. Freddie does not flip the coin. Magaly flips the coin. Is the coin still heads up? Note that "flip" here means "reverse".}\\
\\
|step|subquestion|process|result|\\
|---|---|---|---|\\
|1|Is the coin heads up?|Dorian flips the coin.|The coin is tails up.|\\
|2|Is the coin heads up?|Mayra flips the coin.|The coin is heads up.|\\
|3|Is the coin heads up?|Freddie does not flip the coin.|The coin is heads up.|\\
|4|Is the coin heads up?|Magaly flips the coin.|The coin is tails up.| \\
Therefore, the answer (Yes or No) is "No".\\

				\bottomrule
			\end{tabular}
		}
		\caption{One-shot sample used on Coin Flip}
		\label{tab:few_shot_coin}
	\end{table*}
 		\begin{table*}[t]
		\centering
		\scalebox{0.6}{
			\begin{tabular}{ m{64em} } 
				\toprule
                One-shot Sample Used on Last Letter\\
				\midrule
				\midrule
				{\bf Take the last letters of each words in \"Lucky Mireya Jj Kc\" and concatenate them.}\\
\\
|step|subquestion|process|result|\\
|---|---|---|---|\\
|1|What is the last letter of "Lucky"?|"Lucky"[-1] = 'y'|answer = 'y'|\\
|1|What is the last letter of "Mireya"?|"Mireya"[-1] = 'a'|answer = 'y' + 'a' = 'ya'|\\
|1|What is the last letter of "Jj"?|"Jj"[-1] = 'j'|answer = 'ya' + 'j' = 'yaj'|\\
|1|What is the last letter of "Kc"?|"Kc"[-1] = 'c'|answer = 'yaj' + 'c' = 'yajc'|\\
Therefore, the answer is "yajc".\\
				\bottomrule
			\end{tabular}
		}
		\caption{One-shot sample used on Last Letter}
		\label{tab:few_shot_last}
	\end{table*}

\end{document}